# Illumination and Temperature-Aware Multispectral Networks for Edge-Computing-Enabled Pedestrian Detection

Yifan Zhuang, Ziyuan Pu, *Member, IEEE*, Jia Hu, *Member, IEEE*, and Yinhai Wang, *Senior Member, IEEE*

*Abstract*—Accurate and efficient pedestrian detection is crucial for the intelligent transportation system regarding pedestrian safety and mobility, e.g., Advanced Driver Assistance Systems, and smart pedestrian crosswalk systems. Among all pedestrian detection methods, vision-based detection method is demonstrated to be the most effective in previous studies. However, the existing vision-based pedestrian detection algorithms still have two limitations that restrict their implementations, those being real-time performance as well as the resistance to the impacts of environmental factors, e.g., low illumination conditions. To address these issues, this study proposes a lightweight Illumination and Temperature-aware Multispectral Network (IT-MN) for accurate and efficient pedestrian detection. The proposed IT-MN is an efficient one-stage detector. For accommodating the impacts of environmental factors and enhancing the sensing accuracy, thermal image data is fused by the proposed IT-MN with visual images to enrich useful information when visual image quality is limited. In addition, an innovative and effective late fusion strategy is also developed to optimize the image fusion performance. To make the proposed model implementable for edge computing, the model quantization is applied to reduce the model size by 75% while shortening the inference time significantly. The proposed algorithm is evaluated by comparing with the selected state-of-the-art algorithms using a public dataset collected by in-vehicle cameras. The results show that the proposed algorithm achieves a low miss rate and inference time at 14.19% and 0.03 seconds per image pair on GPU. Besides, the quantized IT-MN achieves an inference time of 0.21 seconds per image pair on the edge device, which also demonstrates the potentiality of deploying the proposed model on edge devices as a highly efficient pedestrian detection algorithm.

*Index Terms*—Pedestrian Detection, Illumination and Temperature-Aware, Multispectral Networks, Sensor Fusion, Network Quantization

## I. Introduction

PEDESTRIAN detection is a crucial task for intelligent transportation systems associated with pedestrian safety and mobility. In general, detection accuracy and efficiency are two major metrics impacting the effectiveness of the systems. Usually, the control actions of the systems are triggered according to real-time pedestrian detection results, e.g., vehicle control of Advanced Driver Assistance Systems (ADAS) [1], and traffic signal control of smart pedestrian crosswalk systems [2, 3]. Wrong detection results can generate false signals to the control decision modules, and the delay in pedestrian detection can lead to untimely decision-making. Consequently, erroneous control actions are highly potential to result in pedestrian-related crashes. Thus, an accurate and efficient pedestrian detection algorithm is significant for improving pedestrian safety and mobility.

Previously, multiple sensing technologies have been applied for pedestrian detection, including vision-based sensing [4], LiDAR [5], radar-based sensing [6], etc. Among the existing sensing technologies, the vision-based method has been demonstrated to be the most effective for pedestrian detection due to the affluent information, and cost-effective features. [7, 8]. On the methodology side, tons of efforts have been made to improve the accuracy and efficiency of pedestrian detection using video or image data [9, 10, 11, 12]. According to previous studies, one crucial factor limiting the detection performance is the environmental illumination [13, 14, 15]. To mitigate the negative influence of low illumination, thermal image data is utilized as a complementary data source when visual image data quality is impacted by low illumination conditions [16]. Basically, multispectral algorithms have been employed to fuse thermal and visual image data. However, there are still several specific disadvantages existing in the previous studies which considerably restrict the accuracy and efficiency. Firstly, in terms of detection accuracy, most existing studies only consider the illumination factor to compute the fusion weights, but the temperature information extracted from thermal images are not utilized to improve the overall detection accuracy. Besides, dynamic fusion weights were not optimized in previous studies. Secondly, for the algorithm efficiency, even the existing neural network-based algorithms highly improved the detection accuracy, the accompanied additional computational complexity impaired their real-time performance.

Therefore, being motivated by the above considerations, the

Yifan Zhuang is with Department of Civil and Environmental Engineering, University of Washing, Seattle, WA 98195 USA (e-mail: zhuang93@uw.edu).

Ziyuan Pu is with School of Engineering, Monash University. Jalan Lagoon Selatan, 47500 Bandar Sunway, Malaysia (email: ziyuan.pu@monash.edu).

Jia Hu is with Key Laboratory of Road and Traffic Engineering of the Ministry of Education, Tongji University, Shanghai, 201804, China (email: hujia@tongji.edu.cn).

Yinhai Wang is with Department of Civil and Environmental Engineering, University of Washing, Seattle, WA 98195 USA (e-mail: yinhai@uw.edu).

primary objective of this paper is to propose an Illumination and Temperature-Aware Multispectral Network (IT-MN) for multispectral pedestrian detection. The major contributions of this study can be summarized in the following five points:

1) Advancing a one-stage detection network called IT-MN for edge-computing-enabled pedestrian detection by fusing visual and thermal images.

2) Designing a convolutional neural network (CNN)-based fusion weight computation network (FWN) as a component of IT-MN dedicatedly for providing fusion weights from visual and thermal images.

3) Designing a novel late-fusion strategy by utilizing all feature maps for feature extraction and fusion on multiple scales in order to further increase accuracy.

4) Apply network quantization to reduce the model size and shorten the inference time, especially on the edge device that always has weak computing power.

5) Optimize the default box generation by reducing the box number as well as selecting specific box aspect ratios.

The remainder of the paper is organized as follows. Section II reviews previous research work in the topic of multispectral pedestrian detection. Section III presents the details of the proposed IT-MN. Section IV shows the experimental results and discusses the corresponding implications. Finally, the paper summarizes the research findings and future work in Section V.

## II. Literature Review

Most existing multispectral pedestrian detection algorithms were developed based on two published datasets – OSU Color-Thermal Database [17] and KAIST Multispectral Pedestrian Dataset [18]. The major difference is the camera mobility where OSU dataset is collected by fixed cameras and KAIST dataset is collected by moving cameras. Thus, the frequent change of background and illumination in KAIST dataset almost disable the application of conventional object detection methods, such as background subtraction [19], and increase the difficulty of distinguishing the object from the background as well. With the advancement of connected and autonomous vehicles (CAVs) and other IoT implementations, pedestrian detection is no longer satisfied by detecting objects with a fixed background. In order to demonstrate the robustness of the proposed method in a more challenging scenario, this research adopts KAIST dataset for both model training and evaluation. A detailed dataset description presents in Section III.

Multispectral pedestrian detection methods can be divided into two categories – conventional methods and deep learning methods. The advantages of conventional methods include explainable structure and simple deployment, which make it easy for parameter adjustments and model transferring. One fundamental method is proposed by Soonmin, etc [18], who combines the aggregated channel features (ACF) [20] and thermal histogram of oriented gradients (HOG) features [21] for pedestrian detection. They build the baseline of KAIST dataset using the evaluation metric of miss rate (MR). And the baseline MR is 64.76%. Besides, they found that thermal images make less contribution to instant pedestrian detection since the context information in thermal images is not as sufficient as visual images. In addition, the manually designed feature descriptors in conventional methods cannot meet the demand for higher accuracy and robustness in various environments. However, the development of deep learning methods, e.g., CNN, brings a powerful tool to solve this challenging demand. Inspired by other detection work of fusing visual and depth images [22], Jörg Wagner, etc. apply the ACF for region proposal to locate potential targets and then use CNN for classification on these proposed regions, which are called region of interest (RoI) [23]. Compared with the baseline built by Soonmin [18], their work reduced MR significantly from 64.76% to 43.80%, which indicates a great necessity to use deep learning methods for pedestrian detection.

Different improvement has been conducted to increase the accuracy and to reduce MR by using deep learning methods. The most influential factor is optimizing fusion strategies. There are generally three fusion strategies – early fusion, middle fusion, and late fusion. Early fusion indicates stacking visual and thermal images before getting into the network. Middle fusion indicates that the layer concatenation happens in the middle of network. And late fusion indicates that the concatenation is conducted at the last layer. Previous work shows that the late fusion is always better than early fusion. In Jörg's research, MR of early fusion is 53.94% which is almost 10% higher than late fusion [23]. More than early and late fusion, middle fusion is explored as well [24]. Daniel König, etc. build their model based on Faster R-CNN [25] and compare detection results among different fusion strategies [26]. In their results, the middle fusion performs the best while the late fusion is very close. The difference in MR between them is less than 0.1%. And MR of middle and late fusion is 9% lower than early fusion. Furthermore, an attention-based fusion developed from the late fusion is proposed [44]. Instead of concatenating layers directly, Yongtao Zhang and his colleagues utilize both local and global attended decoding for layer fusion and get an impressively low MR as a result.

To further reduce MR, the illumination factor is also taken into consideration. Since the idea of fusing visual and thermal images comes from reducing influences of illumination, the equal contribution of two input sources at the fusion stage may not perform as well as dynamic weights computed by current illumination conditions. Starting from this thought, research work has also been explored in designing illumination-aware (IA) network to compute fusion weights [27]. Dayan Guan, etc. design an IA network which uses the feature map from the concatenation layer of visual and thermal streams [28]. And output weights from IA network are used for fusion of both classifier and localization network. Their experiment results show that the model with IA network greatly reduces MR to 29.62%, which is 3% lower than the model without IA network. But another factor of environment temperature may also generate impacts on the final results via thermal images [29]. Only a few pieces of research consider this temperature factor in the fusion process. Yunfan Chen, etc. fuse visual and thermal features at different levels for region proposal [42].

Although detection accuracy is an important evaluation metric for a model, its efficiency is also meaningful especially in some applications where computing power is limited. Nowadays, as a huge amount of real-time image data is generated by pervasive surveillance cameras in cities, cloud computing may not accommodate the exceptionally demand of computational resources for dealing with such a huge data storage and transferring bandwidth [30]. Thus, it raises the demand of a more efficient pedestrian sensing algorithm to analyze images on the sensor side, which is so-called edge computing [39]. Most state-of-the-art detection models are two-stage models, e.g., Faster R-CNN, in consideration of accuracy. Comparably, one-stage models without the region proposal, e.g., Single Shot MultiBox Detector (SSD) [32], have achieved a better trade-off between accuracy and efficiency. Thus, to shorten the model's inference time, Ya-Li Hou, etc. apply SSD as the base model and fuse visual and thermal streams at different feature levels [31]. Compared with other state-of-the-art models [28], the inference time reduces from 0.25 second per frames to 0.03 second on the same computing platform with GPU, which makes it possible to be deployed on the edge side at real-time speed after model simplification such as quantization [44] and distillation [45].

Besides, some researchers also consider the current status of lacking thermal camera devices. Thus, they train models using both visual and thermal images, but only use the visual image as input on the testing or deployment stage [34]. This method could be taken as an emergency tool when the thermal camera does not exist or work. In contrast to the prior method, Yanpeng Cao, etc. train the model using only visual images, but test or deploy model with both visual and thermal images [35]. And this method can solve the common problem of insufficient or missing thermal training data. However, both of them do not utilize thermal information on the training or testing stage. It is reasonable that both of them have higher MR than other models.

III. METHODOLOGY

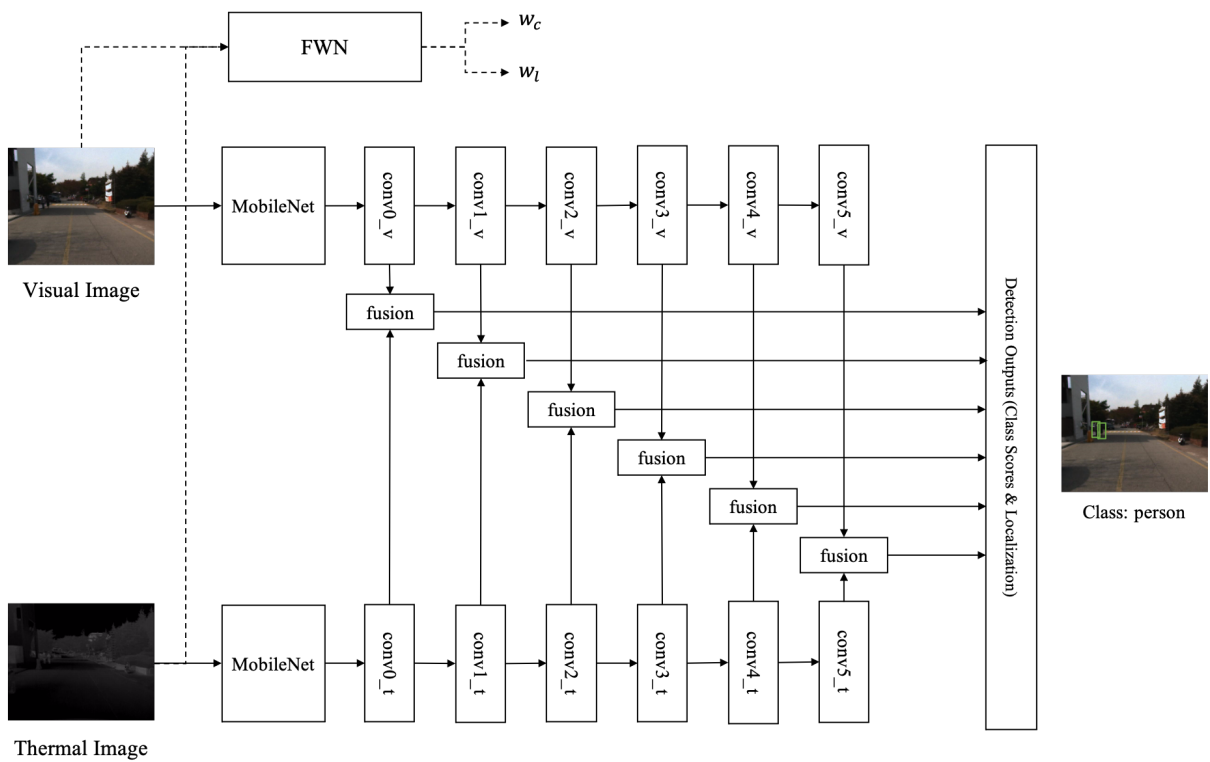

Fig. 1 Framework of IT-MN

The framework of IT-MN is built on SSD. The introduction of IT-MN starts with a brief explanation of the single-stream SSD and two-stream SSD. The single-stream SSD [32] is called S-SSD and the dual-stream SSD is D-SSD. Different from the S-SSD, there are two input streams of visual and thermal images for D-SSD. And the feature extractor of D-SSD is built on MobileNet [43], instead of VGG-16 [40] and ResNet-50 [37], by removing the last average pooling and fully connected layers. Comparing with VGG-16 and ResNet-50, MobileNet is more efficient with much fewer parameters. The parameter number of MobileNet is only 1/8 of ResNet-50 and 1/39 of VGG-16. In this paper, MobileNet v2 is used which absorbs advantages from ResNet. Benefit from both residual block and separable convolution, MobileNet has a strong feature extraction ability and a low computing complexity. Thus, MobileNet is selected as the backbone for D-SSD. The fusion strategy is late fusion, and fusion weights $w_c$ and $w_l$ in Figure 1 are computed through the FWN, where $w_c$ is the weight to fuse the classification score vector, and $w_l$ is the weight to fuse localization coordinate vector. The fusion net is shown in Figure 1, whose function is fusing feature maps from two streams by pointwise addition, and then conducting the class prediction and coordinate regression. The detailed structure of the fusion net will be described in Section III (A). And the default box generation will be discussed in Section III (B). The FWN is applied to promote the detection accuracy under various illumination and temperature conditions, which will be discussed in Section III (C). And the network quantization will be discussed in Section III (D). In Figure 1 and the following figures, the abbreviation "conv" represents the convolutional layer. And activation layers and batch normalization layers are ignored. In Figure 1, there are 6 feature maps used for class prediction and bounding box regression from the backbone, which are from conv0 to conv5 layers. The feature map sizes are $38 \times 38$, $19 \times 19$, $10 \times 10$, $5 \times 5$, $3 \times 3$, and $1 \times 1$ respectively. Each cell on selected feature maps will generate 4 default boxes for object prediction. The coming four sub-sections will introduce feature fusion, default box generation, FWN and network quantization in order.

*A. Feature Fusion*

Based on previous research work, the late fusion has a better performance than the early fusion [23, 31]. Different from the previous work using two-stage detection network that generates region proposals firstly and then regresses and classifies each proposed region, this paper adopts a one-stage framework D-SSD. Thus, the late fusion may not be the best strategy in this case due to the change of the detection framework. Therefore, it is necessary to evaluate different fusion strategies based on D-SSD backbone. In order to find suitable fusion strategies in the current application, three kinds of fusion strategies are compared – early fusion, middle fusion, and late fusion. These three frameworks are shown in Figure 2.

The early fusion, shown in Figure 2 (a), concatenates original visual and thermal images as the input for the detection network. The middle fusion, shown in Figure 2 (b), concatenate feature maps from two streams after passing through conv3_v/conv3_t. The concatenation process applies Network in Network (NIN) [38], referenced from Daniel's work [26]. The late fusion, in Figure 2 (c), does not concatenate feature maps directly during the feature extraction process. Instead, the fusion is conducted in the class prediction and bounding box regression process. In order to highlight the differences among these three strategies, only convolutional layers are shown in Figure 2 from conv0 to conv5, while other layers, e.g., ReLU and Batch Normalization, are omitted.

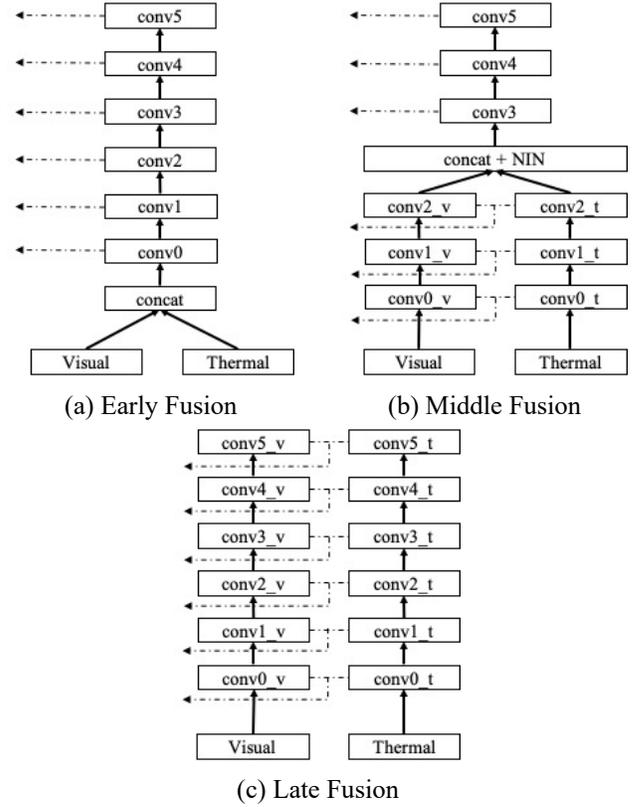

Fig. 2 Frameworks of Fusion Strategies

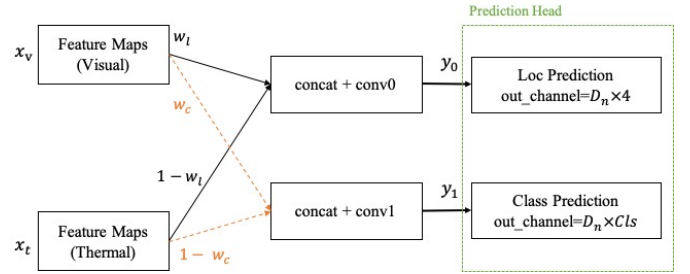

Fig. 3 Localization and Classification Fusion Framework

Different from early fusion that concatenates input images directly, the middle and late fusion strategies will fuse feature maps or predictions after passing through several convolutional layers. The difference between middle and late fusion is that middle fusion conducts the feature extraction separately in two streams on low feature levels before fusion, while late fusion conducts separate feature extractions in two streams on all feature levels. From conv0 to conv3 in middle fusion and conv0 to conv5 in late fusion, the fusion process on a pair of feature maps at the same level keeps the same, whose framework is shown in Figure 3. In Figure 3, $x_v$ and $x_t$ indicate feature maps from visual and thermal streams respectively. $y_0$ and $y_1$ are output of feature fusion, which are also the input of prediction head. $Cls$ indicates the object class number, and $Dn$ indicates the default box number. The input is two feature maps from visual and thermal streams respectively. The fusion weights $w_c$ and $w_l$ computed by FWN are applied to compute weighted sum of classification scores and bounding box coordinates from visual and thermal channels. $w_c$ is the weight to fuse feature maps used for classification, and $w_l$ is the weight to fuse feature maps used for bounding box prediction. The fusion operation is a pixel-wise addition. Then the fused feature map will pass through a convolutional layer for better fusion. As a result, the final fused results are computed through Equation (1) and (2), where $f_0$ and $f_1$ represents conv0 and conv1 respectively.

$$y_0 = f_0(w_l \cdot x_v + (1 - w_l) \cdot x_t) \quad (1)$$

$$y_1 = f_1(w_c \cdot x_v + (1 - w_c) \cdot x_t) \quad (2)$$

*B. Default Box Generation*

This IT-MN is specifically designed for efficient multispectral pedestrian detection. In order to reduce the inference time, one method is reducing the number of default boxes without modifying major network structure. Another possible method is reducing model precision, which will be discussed in Section III (D). However, the first method will decrease the detection accuracy, because the basic idea of proposing RoI is a dense selection. More default boxes indicate a higher probability to cover target objects. In the standard SSD network, each pixel of the feature map will generate multiple default boxes with different aspect ratios. The number could be 4 or 6. The default box example is shown in Figure 4, where $L$ and $W$ represent one default box's length and width respectively. The box aspect ratio $a_r$, whose definition is shown in Equation (3), includes 1, 2, 3, 1/2, 1/3. On a 4 × 4 example feature map, the aspect ratios applied are 1, 2 and 1/2 and the total number is 4. And different colors represent aspect ratios.

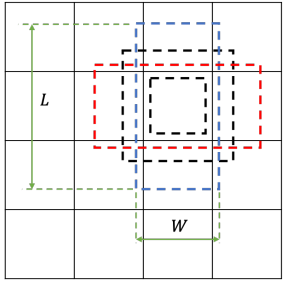

Fig. 4 Default Boxes

In most scenarios of pedestrian detection, the aspect ratio of a pedestrian is always less than one. In addition, the original video frame aspect ratio is greater than one. After resizing the frame shape to a square, the aspect ratio of pedestrian will further reduce. An example scenario is shown in Figure 5, where part (a) shows the original image and part (b) shows the resized image as the input of IT-MN. Thus, those bounding boxes, whose aspect ratio is greater than one, do not have a significant contribution to final detection results. Based on this fact, the aspect ratios in IT-MN are 1, 1/2, 1/3. After setting aspect ratios, the actual size of bounding boxes is referenced to the original work of S-SSD [32]. Suppose there are m feature maps used for detection, the box scale $s_k$ for $k^{th}$ feature map is shown in Equation (4), where $s_{max}$ and $s_{min}$ are set as 0.9 and 0.2 respectively. And $L_k$ and $W_k$ are computed using Equation (5) and (6).

The number of default boxes on all feature levels is set to 4 with optimized aspect ratios. Therefore, the original default box number is 8,732 and the current default box number is 5,820 with a decrease of 33%, which contributes to the decrease of the inference time. The improvement in speed and influence on detection accuracy will be evaluated in Section IV (C). And the evaluation results prove that this number reduction has little negative impact on the accuracy.

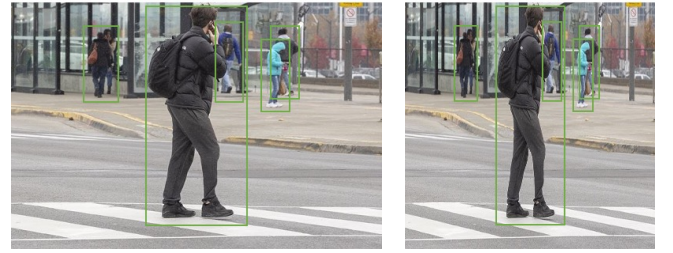

(a) Orginal Image    (b) Resized Image

Fig. 5 Pedestrian Sample

$$a_r = W/L \quad (3)$$

$$s_k = s_{min} + \frac{s_{max} - s_{min}}{m - 1}(k - 1) \quad (4)$$

$$L_k = s_k\sqrt{a_r} \quad (5)$$

$$W_k = s_k/\sqrt{a_r} \quad (6)$$

*C. Fusion Weight Computation Network*

Previous work considers the influences of illumination on detection results and thus designs IA network [27, 28]. In most cases, detection using the visual stream works better in bright conditions, while detection using the thermal stream works better in dark conditions. The input of the IA network is the visual image to estimate the illumination impacts. In addition to illumination impacts, the temperature will affect the detection results via the thermal stream as well. When the temperature is high, it becomes hard to distinguish the background and pedestrians, because their heat radiation is similar. In this situation, the visual stream should have a larger weight in the fusion process. However, the IA network is not capable of analyzing temperature impacts. It is necessary to introduce the temperature awareness into the network. Therefore, this paper proposes a so-called FWN to compute the fusion weights with considering the impacts of both factors. The framework of FWN is shown in Figure 6, and its relationship with other modules in IT-MN is shown in Figure 1. The framework of FWN is referenced from VGG-16, but it is simplified by removing partial convolutional layers and replacing all general convolution with separable convolution. In Figure 6, concatenate operation is working on channel dimension. fc indicates the fully connected layer. Following Figure 1, all ReLU and batch normalization layers are omitted. The parameters of convolutional and fully connected layers in FWN, including input channel number, output channel number and kernel size, are shown in Table 1. The pooling operation is max pooling with downsampling factor of 2. And a sigmoid function is added to the end of the fc layer. The inputs of FWN are visual and thermal images. And the output is synthesized illumination-temperature parameters $w_c$ and $w_l$, which are fusion weights for classification and localization respectively. Considering FWN is also a trainable network, this module does not need a separate training, which indicates the whole training process is an end-to-end training.

TABLE 1 Framework Parameters of FWN

| Layers | In Channel | Out Channel | Kernel Size |
| --- | --- | --- | --- |
| conv0 | 3 | 16 | 3 |
| conv1 | 32 | 64 | 3 |
| conv2 | 64 | 128 | 3 |
| conv3 | 128 | 256 | 3 |
| conv4 | 256 | 128 | 3 |
| conv5 | 128 | 64 | 3 |
| fc | 64 | 2 | - |

The proposed FWN extracts features from both visible and thermal images to generate better fusion weights than previous works by using single visible image to generate fusion weights or applying equivalent weights. In order to analyze the FWN qualitatively, feature maps from layer "conv5" are visualized in Figure 7 and Figure 8. In both figures, part (a) shows the feature map from single visible image, and part (b) shows the feature map from visible and thermal images. The input image pair is shown in part (c) and part (d), where green bounding boxes mark the person object to be detected. Figure 7 and Figure 8 show results from daytime and nighttime correspondingly. For example, the highlight area in Figure 7 (a) and (b) refers to the attention location. Comparing these two feature maps, it can be found that Figure 7 (b) only focuses on the person object after adding thermal images as input. Additionally, Figure 7 (a) also pays attention to the vehicle object on the left side, which is irrelevant in this task. Similar qualitative analysis results can be concluded from Figure 8. After adding the thermal image, the output feature map focuses on the person object. Figure 8 (a) shows that it focuses more on area with texture information. According to the visualization, the proposed FWN can help the detection model better concentrate on the detection target.

*D. Network Quantization*

The computing power on edge devices, e.g., Jetson Nano and Raspberry Pi, is always limited compared to PC or server. In order to implement deep learning methods on these devices, the model simplification is important. The final target is reducing the computing complexity and avoiding losing much accuracy. In this paper, model quantization is applied, whose basic idea is transforming model weights from high-precision format to low-precision format, e.g., from 32-bit floating point to 8-bit integer [44, 46]. The advantage of model quantization is not modifying the model structure but reducing the computing complexity significantly though the number of parameters does not change. Take the above transformation from 32-bit floating point to 8-bit integer as an example, the model size and computing complexity will reduce by 75% with the same number of model parameters. As the result, the overall inference time will reduce by more than 50% in theory in consideration of data loading time. The general quantization process follows Equation (7), where $q$ is the quantized value, $x$ is the original value, $s$ is the scale, $z$ is the zero point and $Q$ is the quantization function. In this paper, the quantization tool from PyTorch is applied. To keep as high accuracy as possible after model quantization, a post-training using the quantized model for a few epochs to calibrate the model weights on the same training dataset is necessary.

$$q = Q(x) = \left\lfloor \frac{x}{s} + z \right\rfloor \qquad (7)$$

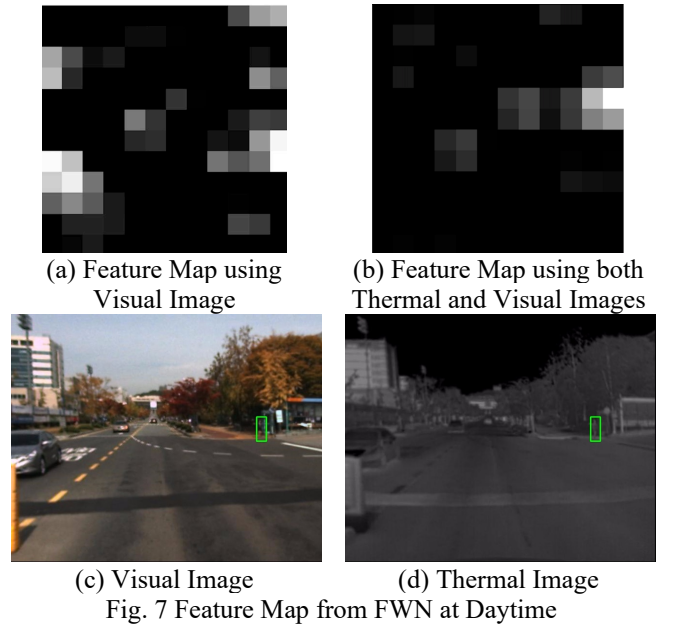

(a) Feature Map using Visual Image  (b) Feature Map using both Thermal and Visual Images

(c) Visual Image  (d) Thermal Image

Fig. 7 Feature Map from FWN at Daytime

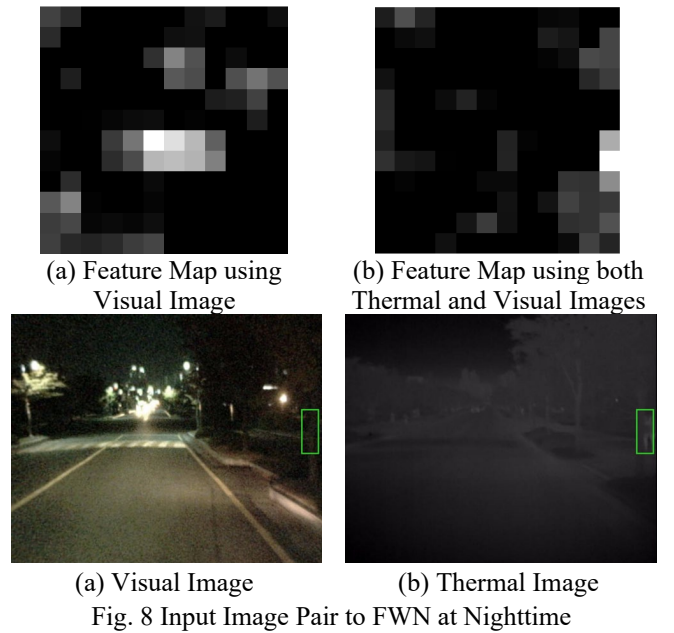

(a) Feature Map using Visual Image  (b) Feature Map using both Thermal and Visual Images

(a) Visual Image  (b) Thermal Image

Fig. 8 Input Image Pair to FWN at Nighttime

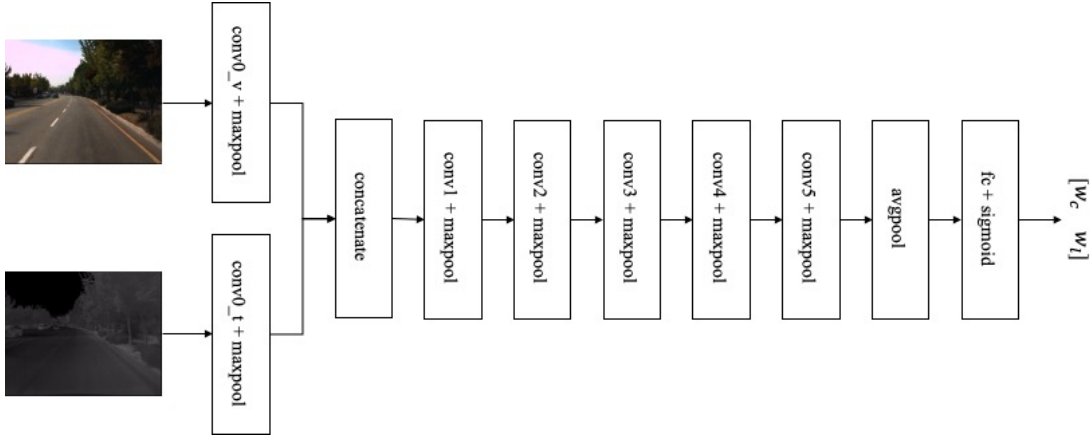

## IV. EXPERIMENTAL DESIGN AND RESULTS

### A. Experimental Design

Experiments are conducted on the KAIST dataset [18], which was published in 2015. The training and testing images were collected from a pair of in-vehicle visual and thermal cameras. Compared to the datasets collecting by the camera at a fixed location, this dataset is more challenging since the background and environment illumination change frequently. As a result, some effective methods cannot be applied in this scenario such as background subtraction. The images in KAIST dataset are taken at 20Hz with a resolution of $640 \times 512$. Image registration of visual and thermal cameras has been completed in this dataset. Thus, each pair of visual and thermal images can be considered to have no perspective difference. And all images are manually labeled with over 100,000 dense annotations in total. There are six scenarios collected in this dataset, which are campus, road and downtown on both daytime and nighttime. Each scenario has both training and testing sub-datasets. The training and testing are conducted on all scenarios. Considering that images are collected continuously using an on-board camera, the adjacent frames are similar. And partial images do not contain objects or have uncleared labels such as "person?". As suggested in previous work [26, 41, 47], the data sampling and cleaning are necessary to filter out most frames containing no target object, which will lead to an unbalanced data distribution, and make network more likely to predict the classes with more training data. In order to train the model more efficiently, data augmentation is applied, which includes random sized cropping, horizontal mirror and saturation change, etc. All input images are resized to $300 \times 300$.

In this paper, experiments were conducted using the PyTorch as the deep learning framework on PC equipped with GPU of Nvidia TITAN XP. Besides, a Raspberry Pi 4 [48] is used as the edge platform to evaluate model efficiency on the edge side. The backbone of IT-MN is pre-trained on ImageNet [33]. The initial weights of additional layers are randomly set following a uniform distribution. The batch size used in the training process is 8. Due to the limitation of GPU memory, the model weights will be updated every four-batches. As a result, the equivalent batch size becomes 32. On the testing stage, the batch size becomes 1 to mimic the image sequence in the practice. The optimization method is stochastic gradient descent. The initial learning rate is 0.001 with a learning rate adjustment method of cosine decay. And the total training epoch number is 200. To improve the training process, the focal loss [49] is applied as the improvement of general cross-entropy loss, whose definition is shown in Equation (8), where $\gamma$ is the hyperparameter. When using cross-entropy loss, training may become insufficient since most samples are easy negatives. This makes it difficult for a model to learn rich semantic information. Comparably, focal loss can reduce this negative impact.

$$FL(p_t) = -(1 - p_t)^\gamma \log(p_t) \qquad (8)$$

To evaluate the detection results, the log-average MR is adopted for accuracy evaluation which is shown in Equation (9). MR reflects missing object number among all positive objects, which is more meaningful than the average precision in the transportation field since the missing detection may lead to serious transportation safety issues.

$$MR = \frac{FN}{TP + FN} \qquad (9)$$

where FN indicates the number of false negatives, and TP indicates the number of true positives. A predicting bounding box is considered as a true positive when it matches a ground truth box with intersection of union (IoU) greater than 0.5 [18]. Since there will be multiple bounding boxes matching a single ground truth, only the box with the highest IoU will be the true positive. Unmatched predicting and ground truth boxes are considered as false positives and false negatives respectively. The log-average MR is computed by averaging MR at nine false positives per image (FPPI) rates evenly spaced in a log-space from 0.01 to 1 [4]. In addition, the model inference is also an important factor. To apply these algorithms on the edge side, efficiency is important. The metric of computing the inference time (second per frame) is adopted to evaluate the algorithm efficiency.

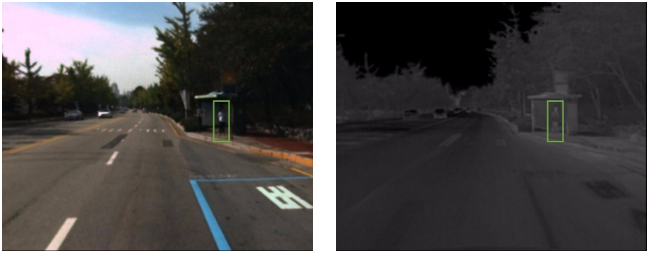
(a) Day Road (Visual)  (b) Day Road (Thermal)

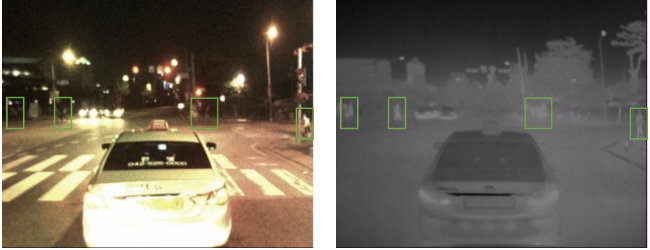
(c) Night Road (Visual)  (d) Night Road (Thermal)

Fig. 9 Positive Detection Samples in Road Scenario

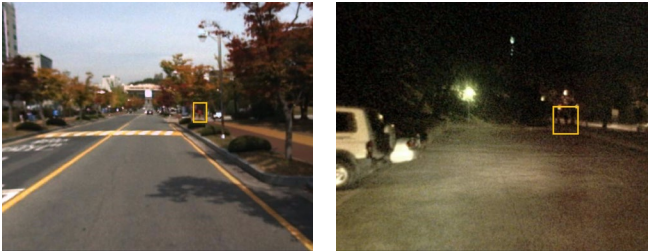
(a) Day Campus  (b) Night Campus

Fig. 10 Missing Detection Samples in Campus Scenario

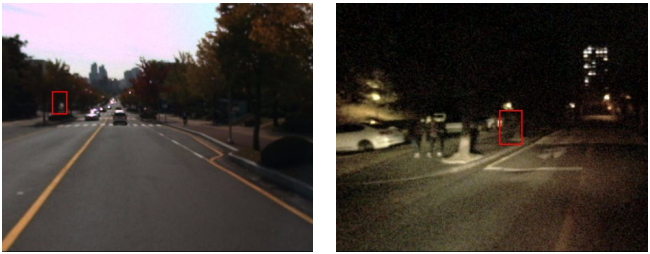
(a) Day Campus  (b) Night Campus

Fig. 11 False Detection Samples in Campus Scenario

The experiment results are shown in the following sections. Section IV (B) is comparing the accuracy of the proposed algorithm with the selected baseline models in terms of MR and inference time. And the log-log plots of MR against FPPI are presented in this part. In order to show the improvement by fusing the RGB and thermal images with the FWN, Section IV (C) presents the ablation tests of using different streams, fusion strategies, and awareness networks. And Section IV (D) shows the improvements by optimizing the default box generation. All efficiency tests are conducted on both PC and Raspberry Pi sides.

*B. Evaluating the Detection Accuracy of the Proposed IT-MN*

In this section, the proposed algorithm is evaluated based on the comparison with state-of-the-art multispectral pedestrian detection methods in terms of MR and inference time. The comparison models are Late-fusion SSD (L-SSD) [31], CWF-CNN [13], IATDNN [28] and MLF-CNN [42]. Six scenarios in KAIST dataset are merged into three groups based on the collection time – day, night and all time. The comparison results of MR are shown in both Figure 12 and Table 2. Figure 12 plots MR versus FPPI and Table 2 presents the mean MR in terms of all day, daytime and nighttime.

In Table 2, both mean MR and average precision (AP) are computed. Comparing two SSD-based algorithms (L-SSD and IT-MN), the proposed model reduces MR significantly from 43% to 16%, which proves the necessity of applying the extra network to compute the fusion weights instead of fusing feature maps with equal contribution. And IT-MN has a much lower MR than other state-of-the-art algorithms including CWF-CNN, IATDNN and MLF-CNN. The mean MR of IT-MN is around 14%, and other algorithms have at least 10% higher than the proposed IT-MN. Meanwhile, the quantized IT-MN, which is designed for the edge device, is also evaluated. The only difference between the quantized and original IT-MN is the quantization operation on model weights from 32-bit floating point to 8-bit integer format. The quantized IT-MN is called "Quant. IT-MN" in the following part. Contributed by the post-training process using the same training dataset, the MR of Quant IT-MN is about 0.3% higher than IT-MN. In Table 2, comparing experiment results using MR and AP, they are consistent. In the following sections, only MR is applied for evaluation.

TABLE 2 Comparison of Detection Accuracy in terms of All Day, Daytime and Nighttime

| Algorithms | MR (All) | MR (Day) | MR (Night) | AP (All) | AP (Day) | AP (Night) |
|---|---|---|---|---|---|---|
| L-SSD | 43.06% | 50.73% | 35.38% | 50.61% | 52.02% | 48.77% |
| CWF-CNN | 31.36% | 31.79% | 30.82% | 65.36% | 66.51% | 64.59% |
| IATDNN | 29.62% | 30.30% | 26.88% | 67.25% | 68.33% | 67.02% |
| MLF-CNN | 25.65% | 25.22% | 26.60% | 68.22% | 68.99% | 67.60% |
| IT-MN | **14.19%** | **14.30%** | **13.98%** | **73.22%** | **72.80%** | **73.51%** |
| Quant. IT-MN | 14.55% | 14.67% | 14.29% | 71.93% | 71.10% | 72.20% |

In Figure 12, the x-axis is the false positive number per image (FPPI) and the y-axis is MR. FPPI ranges from 0.001 to 1.0 using a log scale and the mean value of MR is computed in the range between 0.01 and 1.0. Curves under three conditions are plotted separately as shown from Figure 12 (a) to (c). According to the plots, the curve of IT-MN is lower than the curves of all other algorithms, which indicates that IT-MN has lower MR when using different thresholds of detection confidence under all light conditions.

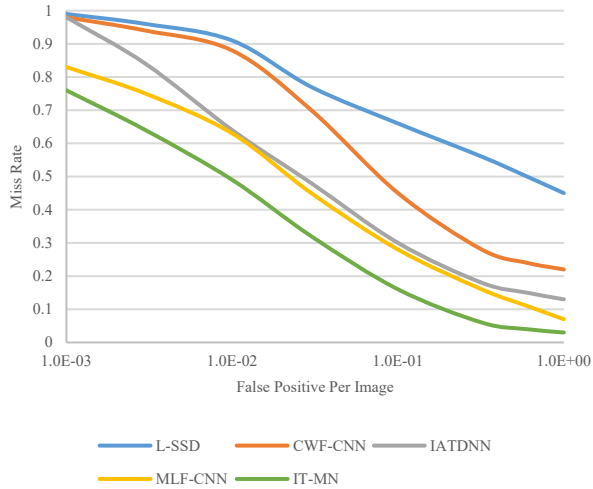

(a) Reasonable All Day

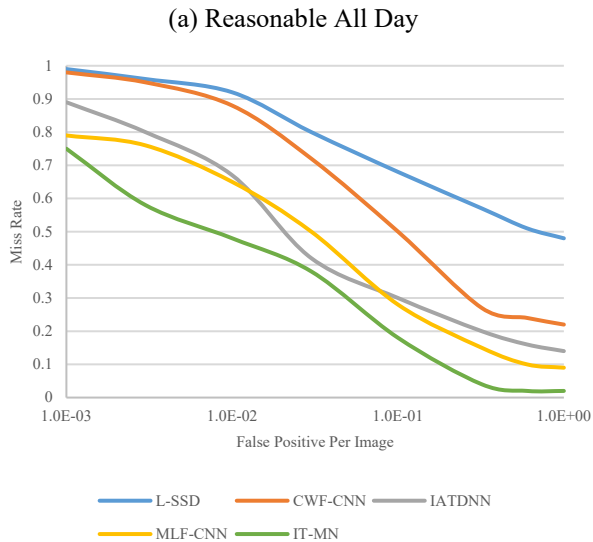

(b) Reasonable Day

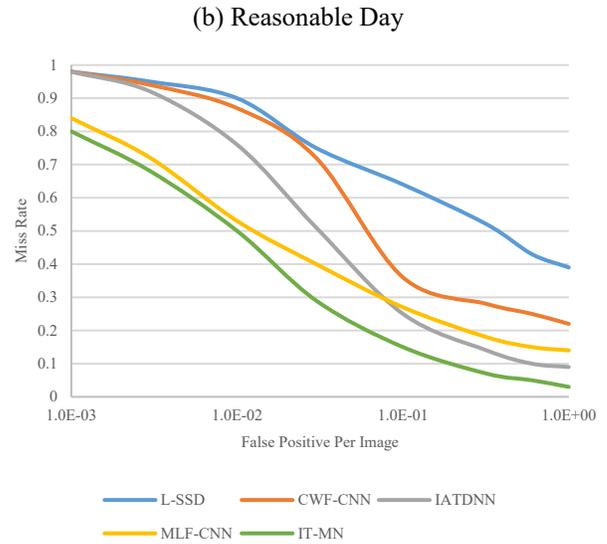

(c) Reasonable Night

Fig. 12 Comparison of Detection Accuracy (MR versus FPPI) in terms of All Day, Daytime and Nighttime

Besides the detection accuracy, the inference time is another important evaluation metric. In this study, the inference time of the proposed algorithm and other baseline algorithms is compared. To demonstrate the potential as an edge computing algorithm, the experiment is conducted separately on GPU and the Raspberry Pi 4 which is popular open-source hardware for edge computing. According to the comparison results shown in Table 4, it can be found that IT-MN has an impressive inference time with 0.03 seconds per frame when using GPU. It is notable that L-SDD achieves the same inference time as the proposed IT-MN, however, which the accuracy is much worse than IT-MN. Compared to other algorithms with similar MR, IT-MN is much faster than others. In particular, the proposed IT-MN is 5 times faster than MLF-CNN when using GPU. For the results using the Raspberry Pi 4, it is similar to the case using GPU that the proposed IT-MN outperforms almost all other algorithms except for L-SDD. In addition, after quantizing the proposed IT-MN, the inference time is improved a lot which is about two times faster than the original IT-MN. In summary, the proposed IT-MN reaches 0.03 and 0.21 seconds per frame, respectively, when using GPU and the Raspberry Pi 4, while other state-of-the-art algorithms with similar MR cost several seconds for processing one image pair. Seen from the absolute numbers of the inference time, the results indicate IT-MN can process dozens of image pairs using GPU. Even for running on the Raspberry Pi 4 which computational power is way more limited than GPU, IT-MN also can process around 5 image pairs per second. Besides using inference time for evaluation, the number of parameters is another indirect evaluation metric. In general, more parameters refer to higher model complexity. Except the Quant. IT-MN, the precision of other models is 32-bit floating point. Comparing the first five models, SSD-based models, i.e., L-SSD and IT-MN, have much fewer parameters than other models. Although IT-MN and Quantized IT-MN have the same number of parameters, the quantized one has a lower precision. Quant. IT-MN has a lower computing complexity than IT-MN. The result indicates the inference time of the proposed IT-MN has a great potential to achieve satisfactory real-time performance for detecting pedestrians through a live stream after a further engineering optimization, e.g., implementation under the C++ framework.

TABLE 4 Comparison of Inference Time on GPU and Raspberry Pi (Unit: Second per Frame)

| Algorithms | Speed (GPU) | Speed (Edge) |
|---|---|---|
| L-SSD | **0.03** [1] | 0.38 [2] |
| CWF-CNN | 0.50 [1] | 5.30 [2] |
| IATDNN | 0.25 [1] | 2.60 [2] |
| MLF-CNN | 0.15 [1] | 1.60 [2] |
| IT-MN | **0.03** [1] | 0.40 [2] |
| Quant. IT-MN | - | **0.21** [2] |

(1) NVIDIA TITAN XP
(2) Raspberry Pi 4

TABLE 5 Comparison of Model Parameters

| Algorithms | Model Parameters |
|---|---|
| L-SSD | 7.1M |
| CWF-CNN | 85.5M |
| IATDNN | 71.2M |
| MLF-CNN | 41.7M |
| IT-MN | 9.6M |
| Quant. IT-MN | 9.6M |

### C. Evaluating Fusion Strategies and Awareness Network

This section will explore the improvement by using different fusion strategies. Conventional methods always rely on visual texture. However, the proposed method relies on both visual and thermal textures. It is important to evaluate different fusion strategies. The first experiment is focusing on evaluating the improvement after adding the thermal images. And the second one is comparing the accuracy with different fusion positions. The third one is comparing the accuracy of different awareness networks, which is used to compute the fusion weights.

The thermal image provides more pedestrian feature information when the illumination condition is poor and visual information is insufficient. In this situation, the pedestrian is easier to be recognized using thermal images. However, the thermal image is sensitive to both environment and pedestrian temperature. When the environment temperature rises or pedestrians are distant from the camera, the difficulty of recognition using thermal image increases as well. Compared with the night scenario, the temperature at daytime is higher, which indicates that other objects besides pedestrians are more recognized in thermal images. It thus increases the difficulty to distinguish pedestrians from the background. Therefore, the thermal image still has some limitations and can be used to improve detection results under certain scenarios. The experiment is conducted to evaluate how different input sources affect performance. Table 6 shows MR comparison results among three networks – S-SSD using the visual image, S-SSD using the thermal image, and IT-MN fusing visual and thermal images.

TABLE 6 Comparison of Different Data Inputs

| Algorithms | MR (All) | MR (Day) | MR (Night) |
|---|---|---|---|
| Visual | 41.27% | 36.79% | 46.22% |
| Thermal | 40.73% | 47.77% | 32.29% |
| Fusion | **14.19%** | **14.30%** | **13.98%** |

According to the results in Table 6, it can be observed that S-SSD using visual image has lower MR during daytime and higher MR during nighttime. Considering that the pixel value of thermal images represents the heat received by the sensor. The higher pixel value means higher temperature. Thus, comparing the average pixel value of images from day and night can obtain the average heat or temperature at different times. The average pixel value of day thermal images is 73% higher than night images, which indicates that temperature at day is higher. Combining this analysis with experiment results from Table 6, thermal images are more suitable for pedestrian detection in the condition of low temperature and illumination. When the illumination is bright enough or the environment temperature is high, the visual image can provide more detailed and reliable information.

The thermal image can improve MR performance at night. However, it may generate negative impacts on MR performance during daytime if it is directly combined with the visual image as the input source without any fusion strategy. In this part, the experiment is conducted on comparing different fusion strategies, including early fusion, middle fusion, and late fusion. These three strategies are elaborated in Section III (A). Table 7 shows the experiment results. The early fusion has the highest MR, which indicates that stacking visual and thermal images directly as one input for the network could not make full use of thermal information. And the late fusion strategy has lower MR than middle fusion, which indicates that weighted fusion on each feature map can be more beneficial.

TABLE 7 Comparison of Different Fusion Strategies

| Algorithms | MR (All) | MR (Day) | MR (Night) |
|---|---|---|---|
| Early Fusion | 29.51% | 29.82% | 28.60% |
| Middle Fusion | 24.77% | 26.60% | 22.33% |
| Late Fusion | **14.19%** | **14.30%** | **13.98%** |

One important contribution of this work is proposing a novel FWN with the consideration of both illumination and temperature factors. In order to evaluate the designed FWN, four kinds of frameworks with different awareness networks are compared – network without awareness network (N-MN), network with IA network (I-MN), network with the temperature-aware (TA) network (T-MN), and the proposed IT-MN. The fusion weight in N-MN is 0.5, which indicates that the final detection results is calculated by averaging the outputs of visual and thermal image data streams. The comparison results of four strategies are shown in Table 8. According to the results, both IA and TA can effectively reduce MR. However, the decreasing scale of these networks is different under different conditions. I-MN has a lower MR in the daytime while T-MN has a lower MR at nighttime. In general, it can be found that IA network has a lower MR on all scenarios than TA network, which indicates that the illumination factor is more apparent than the temperature factor. After integrating IA and TA, the MR is reduced by IT-MN with 20% compared to N-MN.

TABLE 8 Comparison of Different Awareness Networks for Full KAIST Dataset

| Algorithms | MR (All) | MR (Day) | MR (Night) |
|---|---|---|---|
| N-MN | 35.28% | 35.66% | 34.30% |
| I-MN | 27.93% | 27.45% | 28.70% |
| T-MN | 29.65% | 30.80% | 27.61% |
| IT-MN | **14.19%** | **14.30%** | **13.98%** |

In order to further demonstrate the contribution TA network to the detection accuracy, a set of images are selected to test the performance of the algorithms in the environment with low illumination and high temperature. The image pairs with the illuminance lower than the threshold $T_{ill}$ and the temperature higher than the threshold $T_{tem}$. The measurement of illuminance is the sum of L channel pixel value after transforming the visual image into LAB color space. The temperature is measured through the sum of one channel pixel value from the thermal image. Two thresholds are set based on average illuminance and temperature of all training and testing datasets. The experiment results are shown in Table 9. The network with TA has a great improvement in the specific temperature dataset than in the full KAIST dataset, which the MR decreases 5.63% by integrating TA with N-MN in the full KAIST data while the MR decreases 8.09% in the selected image pairs. Furthermore, IT-MN has an outstanding performance in both sets, whose MR is at least 10% lower than other networks. Thus, this comparison proves that the temperature factor can contribute more to the situations with high environmental temperature. As a result, the FWN can generates more benefits for pedestrian detection under various illumination environments, compared with single awareness networks, e.g., IA and TA networks.

TABLE 9 Comparison of Different Awareness Networks for Specific Temperature Dataset

| Algorithms | MR (Night) |
|---|---|
| N-MN | 39.18% |
| I-MN | 33.37% |
| T-MN | 31.09% |
| IT-MN | **20.29%** |

D. *Evaluating the Performance of Default Box Optimization*

Besides proposing the FWN to compute the fusion weights, another framework improvement is optimizing the default box generation by reducing the number of boxes and changing the box aspect ratio. According to Section III (B), the total number of default box decreases by 33% from 8,732 to 5,820. In general, denser default boxes indicate higher detection accuracy but lower speed. Considering the specific detection scenario and aspect ratio of pedestrians, the simplification is expected to have little negative impact on detection accuracy while reducing the inference time. The comparison results in Table 10 and 11 match this expectation. Thus, it achieves a trade-off between accuracy and speed. The optimized default box generation can keep the accuracy at a reasonable level while reducing inference time dramatically by over 25%.

TABLE 10 Comparison of Different Default Box Setting in Accuracy

| Algorithms | MR (All) | MR (Day) | MR (Night) |
|---|---|---|---|
| Improved | 14.19% | 14.30% | 13.98% |
| Original | **14.01%** | **14.10%** | **13.87%** |

TABLE 11 Comparison of Different Default Box Setting in Inference Time (Unit: Second per Frame)

| Algorithms | Speed (GPU) | Speed (Edge) |
|---|---|---|
| Improved | **0.03** [1] | **0.40** [2] |
| Original | 0.04 [1] | 0.60 [2] |

(1) NVIDIA TITAN XP
(2) Raspberry Pi 4

This experiment is conducted on three scenarios, including all day, nighttime and daytime, and two platforms, including

Nvidia Titan XP and Raspberry Pi 4. In Table 10, results show that the network with the original default box setting has a lower MR, since a higher box number can increase the probability to cover objects to be detected. However, these two settings have little difference between MR, and the difference is 0.2%. This difference can be even ignored in practical use. Different from MR, the inference time of two settings has a larger gap as shown in Table 11. IT-MN with an improved setting is 25% faster than the original one. The inference time is 0.03 seconds on the GPU device and 0.4 seconds on the edge device. Considering that the computation power on the edge side is limited, every effort on promoting inference time is very important. In conclusion, the effort by optimizing the default box generation improves the efficiency significantly while having slightly negative impacts on accuracy.

V. CONCLUSION

This paper proposes an efficient and accurate multispectral pedestrian detection algorithm, called IT-MN, considering both illumination and temperature factors. The algorithm is built on SSD with late-fusion strategy and additional FWN to compute the fusion weights. The default box generation is optimized by reducing the proposed box number to shorten the inference time. To further increase the model efficiency on the edge device, the model quantization is discussed and evaluated. The experiment results show that the proposed IT-MN outperforms most state-of-the-art models and achieving a low miss rate (MR) at 14.19%. And this result is also very close to the currently lowest MR at 11.43%. More importantly, the proposed IT-MN can process an image pair every 0.03 seconds using GPU and 0.4 seconds on an edge device without engineering optimization. As the result, its quantized version can reduce this time by almost 50%. In conclusion, IT-MN has achieved excellent performance in both detection accuracy and efficiency. Besides, with characterizing as an illumination and temperature-aware detection algorithm, the proposed algorithm has a great potential for deployment on edge devices for real-time pedestrian detection with the ability to accommodating low illumination conditions.

Future work will be explored in domain adaption which can help bridge the gap between different domains [50, 51, 52, 53]. Different weather will cause the domain shift. Considering that it is impossible to collect and label training data under different weather conditions, e.g., foggy [54] and rainy [55] day, domain adaptive object detection can be a good option to improve the detection performance.

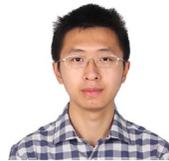
**Yifan Zhuang** received a B.E. degree in Automation from Tsinghua University, Beijing, China (2016) and a M.S. degree in Civil and Environmental Engineering from University of Washington (UW), Seattle, US (2019). Currently, he is working toward the Ph.D. degree in Civil and Environmental Engineering at the UW, Seattle, US. He is currently a Research Assistant at Smart Transportation Application and Research (STAR) Lab of the UW. His major research interests include vision-based object detection, video understanding, advanced sensing technologies, and traffic pattern recognition.

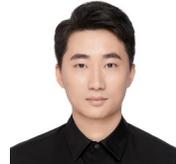
**Dr. Ziyuan Pu** is a Lecturer (Assistant Professor) at Monash University. He received B.S. degree in transportation engineering in 2010 at Southeast University, China. He received M.S. and Ph.D. degree in civil and environmental engineering in 2015 and 2020, respectively, at the University of Washington, US. His research interests include transportation data science, smart transportation infrastructures, connected and autonomous vehicles (CAV), and urban.

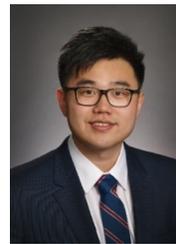
**Dr. Jia Hu** works as a ZhongTe Distinguished Chair in Cooperative Automation in the College of Transportation Engineering at Tongji University. Before joining Tongji, he was a research associate at the Federal Highway Administration, USA (FHWA). He is an Associate Editor of the American Society of Civil Engineers Journal of Transportation Engineering, IEEE Open Journal in Intelligent Transportation Systems, an assistant editor of the Journal of Intelligent Transportation Systems, an advisory editorial board member for the Transportation Research Part C, an associate editor for IEEE Intelligent Vehicles Symposium since 2018, and an associate editor for IEEE Intelligent Transportation Systems Conference since 2019.

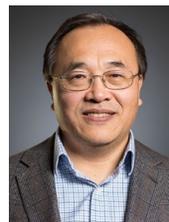
**Dr. Yinhai Wang** is a professor in transportation engineering and the founding director of the Smart Transportation Applications and Research Laboratory (STAR Lab) at the University of Washington (UW). He also serves as the director of USDOT University Transportation Center for Federal Region 10 (PacTrans). He has a Ph.D. in transportation engineering from the University of Tokyo (1998) and a master's degree in computer science from the UW. Dr. Wang's active research fields include traffic sensing, e-science of transportation, transportation safety, etc.